\documentclass[runningheads]{llncs}
\usepackage{latexsym}

\usepackage{times}

\usepackage{rotating} %
\usepackage{soul}
\usepackage{url}
\usepackage[hidelinks]{hyperref}
\usepackage[utf8]{inputenc}
\usepackage{graphicx}
\usepackage[color=blue!10]{todonotes}
\usepackage{amsmath}
\usepackage{booktabs}
\usepackage{algorithm}
\usepackage{algorithmic}
\usepackage{url,ifthen}
\usepackage{amssymb}
\usepackage{comment}
\usepackage{multicol}

\begin{document}

\title{Monte Carlo Graph Coloring}

\author{Tristan Cazenave \and Benjamin Negrevergne \and Florian Sikora}

\institute{Universit\'e Paris-Dauphine, PSL University, CNRS, LAMSADE, 75016 Paris, France \email{\{Tristan.Cazenave,Benjamin.Negrevergne,Florian.Sikora\}@dauphine.fr}}

\maketitle

\begin{abstract}
Graph Coloring is probably one of the most studied and famous problem in graph algorithms. Exact methods fail to solve instances with more than few hundred vertices, therefore, a large number of heuristics have been proposed. Nested Monte Carlo Search (NMCS) and Nested Rollout Policy Adaptation (NRPA) are Monte Carlo search algorithms for single player games. Surprisingly, few work has been dedicated to evaluating Monte Carlo search algorithms to combinatorial graph problems. In this paper we expose how to efficiently apply Monte Carlo search to Graph Coloring and compare this approach to existing ones.
\end{abstract}

\section{Introduction}

Given a graph $G$, a proper coloration of $G$ consists in assigning a color to each vertex of the graph such that no adjacent vertices receive the same color.
The chromatic number $\chi(G)$ of $G$ is the minimum number of colors required to have a proper coloration for $G$. Determining the chromatic number of a graph is probably one of the most studied topics in graph algorithms and discrete mathematics. It has many applications, including scheduling, timetabling,  or communication networks (see references in~\cite{DBLP:journals/itor/MalagutiT10}). Unfortunately, identifying the chromatic number is notoriously hard to solve: it is already NP-hard even if the question is to decide if the graph can be colored with 3 colors,  and it is essentially completely not-approximable~\cite{DBLP:journals/toc/Zuckerman07}.

To cope with this difficulty, the research community has tried a variety of different approaches: mathematical programming~\cite{DBLP:journals/itor/MalagutiT10}, exact moderately exponential algorithms~\cite{DBLP:journals/siamcomp/BjorklundHK09}, approximation algorithms on special graph classes~\cite{DBLP:conf/focs/DemaineHK05}, algorithms of parameterized complexity for structural parameters and data reduction~\cite{DBLP:journals/iandc/JansenK13,DBLP:journals/dam/Cai03a,DBLP:journals/tcs/Marx06a,DBLP:conf/icalp/Lampis18}, heuristics, meta-heuristics, etc. In practice exact methods generally fail to color graphs with more than few hundred vertices~\cite{DBLP:journals/itor/MalagutiT10}, so a large number of publications on graph coloring algorithms focus on the design and improvement of heuristics approaches. 

Early heuristics for graph coloring were often based on pure local search strategies such as {\em TabuSearch}~\cite{hertz1987using}. Nowadays, most efficient modern algorithms are still based a local search strategy, but they combine it with sophisticated exploration techniques to escape local minima (e.g. {\em Variable Neighborhood Search}~\cite{mladenovic1997variable} and {\em Variable Space Search}~\cite{hertz2008variable}). 
Building on a the idea of combining local search with more exploratory search procedures, Fleurent and Ferland have proposed to use the framework of hybrid algorithms which combine a local search operator with a population based algorithm~\cite{fleurent1996genetic}. This idea has inspired a lot of research  in the field (see for example \cite{galinier1999hybrid}), and ultimately led to the state-of-the-art algorithm HEAD~ \cite{moalic2018variations} ({\em Hybrid Evolutionary Algorithm in Duet}).

In comparison with hybrid  algorithms based on local search, very little work has been dedicated to evaluating the performance of Monte-Carlo for graph coloring problem (except \cite{edelkamp2017solving}). This is probably because the idea of discovering highly constrained solutions through random sampling seems counter-intuitive at first. However, modern Monte-Carlo based algorithm  naturally combine random search (which provides exploration) with a tree search driven by a stochastic policy learned  during the search, (which help improving good local solutions).  These two features make modern Monte-Carlo based algorithms good candidates for the graph coloring problem. 

In this paper, we evaluate the performance of two Monte-Carlo based algorithm, {\em Nested Monte Carlo Search} (NMCS)~\cite{CazenaveIJCAI09} and {\em Nested Rollout Policy Adaptation} (NRPA)~\cite{Rosin2011}, for the graph coloring problem. As we will show, our modeling of the coloring problem as a Monte-Carlo search algorithm provides good performance and can compete with state-of-the-art hybrid algorithms which have been studied and improved over the past 30 years.

In Section~\ref{sec:related}, we review related work concerning Monte Carlo Search methods and describe in Section~\ref{sec:mcs} the two we will use in this paper.
In Section~\ref{sec:graph-coloring-as-mc}, we discuss various modeling choices for our approach.
In Section~\ref{sec:other-approaches-for-graph-coloring} we describe the other algorithms for graph coloring that we will compare to ours.
Finally, in Section~\ref{sec:exp}, we conduct thorough experiments to demonstrate the performance of NRPA.

\section{Monte Carlo Search methods and Combinatorial Problems}
\label{sec:related}

Monte Carlo Tree Search algorithms (MCTS) have been most successful in the area of game artificial intelligence~\cite{BrownePWLCRTPSC2012}, and have obtained state-of-the-art in this field. They have also been applied to a variety of other problems in combinatorial optimization problems, but they remain marginally used in this area.  For example, NRPA has been applied to the Traveling Salesman with Time Windows problem \cite{cazenave2012tsptw,edelkamp2013algorithm}, and other applications also deal with 3D Packing with Object Orientation \cite{edelkamp2014monte}, the physical traveling salesman problem \cite{edelkamp2014solving}, the Multiple Sequence Alignment problem \cite{edelkamp2015monte} or Logistics \cite{edelkamp2016monte}.

In 2017, Edelkamp and co-workers have applied Monte Carlo Search to graph coloring~\cite{edelkamp2017solving}. 
 They compare various Monte-Carlo search algorithms such as NMCS, NRPA as well as a SAT-based approach. In their experiments they report that the best results were obtained with NMCS which contrast with our results in this paper. We propose to optimize further the modelling of the problem using node ordering, refined scoring and a different Adapt function. Our optimizations improve much the search time compare to the alternative modellings.

\section{Nested Monte Carlo Search}\label{sec:mcs}

In this section we describe the two Monte Carlo search algorithms that we have considered in the rest of this paper: {\em Nested Monte Carlo Search} (NMCS) and {\em Nested Rollout Policy Adaptation} (NRPA).

As most Monte-Carlo based algorithms, NMCS and NRPA produce a good solution by generating a large number of random sequences of branching decisions (a.k.a moves). The best sequence according to some objective function is then returned as a final solution to the problem.
Since the quality of final sequence directly depends on the quality of the random sequences generated during the search, NMCS and NRPA combine a variety of techniques to improve the quality of the random sequence generator such as tree search, policy adaptation or nested algorithms. 

At the lowest recursive level, the generation of random sequences is driven by a stochastic policy (a probability distribution over the moves). Random sequences are generated based on this policy by sampling moves from the policy using Gibbs sampling, as described in Algorithm~\ref{PLAYOUT}. %
If we have access to background knowledge, it can be encoded as a non-uniform distribution over the moves in the policy. Otherwise, the initial stochastic policy assigns equal probability to each move.  

\begin{algorithm}[tbh]
\begin{algorithmic}
\STATE{playout ($state$, $policy$)}
\STATE{$sequence$ $\leftarrow$ []}
\WHILE{true}
\IF{$state$ is terminal}
\RETURN{(score ($state$), $sequence$)}
\ENDIF
\STATE{$z$ $\leftarrow$ 0.0}
\FOR{$m$ in possible moves for $state$}
\STATE{$z$ $\leftarrow$ $z$ + exp ($policy$ [$m$])}
\ENDFOR
\STATE{choose a move $m$ with probability $\frac{exp (policy [m])}{z}$}
\STATE{$state$ $\leftarrow$ play ($state$, $m$)}
\STATE{$sequence$ $\leftarrow$ $sequence$ + $m$}
\ENDWHILE
\end{algorithmic}
\caption{\label{PLAYOUT}The playout algorithm}
\end{algorithm}

In NMCS, the policy remains the same throughout the execution of the algorithm. However, the policy is combined with a tree search to improve the quality over a simple random sequence generator. At each step, each possible move is evaluated by completing the partial solution into a complete one using moves sampled from the policy. Whichever intermediate move has led to the best completed sequence, is selected and added to the current sequence.  (See Algorithm~\ref{NMCS}.)
The same procedure is repeated to choose the following move, until the sequence has reached a terminal state.

\begin{algorithm}[tbh]
\begin{algorithmic}
\STATE{NMCS ($state$, $level$)}
\IF{level == 0}
\RETURN{playout ($state$, $uniform$)}
\ENDIF
\STATE{$BestSequenceOfLevel$ $\leftarrow$ $\emptyset$}
\WHILE{$state$ is not terminal}
\FOR{$m$ in possible moves for $state$}
\STATE{$s$ $\leftarrow$ play ($state$, m)}
\STATE{NMCS ($s$, $level-1$)}
\STATE{update $BestSequenceOfLevel$}
\ENDFOR
\STATE{$bestMove$ $\leftarrow$ move of the $BestSequenceOfLevel$}
\STATE{$state$ $\leftarrow$ play ($state$, $bestMove$)}
\ENDWHILE
\end{algorithmic}
\caption{\label{NMCS}The NMCS algorithm.}
\end{algorithm}

A major difference between NMCS and NRPA, is  the fact that NRPA uses a stochastic policy that is {\em learned} during the search. At the beginning of the algorithm, the policy is initialized uniformly and later improved using gradient descent steps based the best sequence discovered so far (See. Algorithm~\ref{NRPA}). 
The procedure used to update the policy from a given sequence is given in Algorithm~\ref{ADAPT}. Note that this difference is crucial  because unlike NMCS, NRPA is able to {\em acquire} background knowledge about the problem being solved, and does not require the user to specify it. Ultimately, this knowledge will contribute to speed up the discovery of a good solution.

Finally, both algorithms are nested, meaning that at the lowest recursive level, weak random policies are used to sample a large number of low quality sequences, and produce a search policy of intermediate quality. At the recursive level above, this policy is used to produce sequence of high quality. 
This procedure is applied recursively, in general 4 or 5 times. In both algorithm the recursive level (denoted $l$) is a crucial parameter. Increasing $l$ increases the quality of the final solution at the cost of more CPU time. 
In practice it is generally set to 4 or 5 recursive levels depending on the time budget and the computational resources available.

\begin{algorithm}[tbh]
\begin{algorithmic}
\STATE{NRPA ($level$, $policy$)}
\IF{level == 0}
\RETURN{playout (root, $policy$)}
\ENDIF
\STATE{$bestScore$ $\leftarrow$ $-\infty$}
\FOR{N iterations}
\STATE{(result,new) $\leftarrow$ NRPA($level-1$, $policy$)}
\IF{result $\geq$ bestScore}
\STATE{bestScore $\leftarrow$ result}
\STATE{seq $\leftarrow$ new}
\ENDIF
\STATE{policy $\leftarrow$ Adapt (policy, seq)}
\ENDFOR
\RETURN{(bestScore, seq)}
\end{algorithmic}
\caption{\label{NRPA}The NRPA algorithm.}
\end{algorithm}

\begin{algorithm}[bth]
\begin{algorithmic}
\STATE{Adapt ($policy$, $sequence$)}
\STATE{$polp \leftarrow$ $policy$}
\STATE{$state \leftarrow$ $root$}
\FOR{$move$ in $sequence$}
\STATE{$polp$ [$move$] $\leftarrow$ $polp$ [$move$] + $\alpha$}
\STATE{$z$ $\leftarrow$ 0.0}
\FOR{$m$ in possible moves for $state$}
\STATE{$z$ $\leftarrow$ $z$ + exp ($policy$ [$m$])}
\ENDFOR
\FOR{$m$ in possible moves for $state$}
\STATE{$polp$ [$m$] $\leftarrow$ $polp$ [$m$] - $\alpha * \frac{exp (policy [m])}{z}$}
\ENDFOR
\STATE{$state$ $\leftarrow$ play ($state$, $move$)}
\ENDFOR
\STATE{$policy$ $\leftarrow$ $polp$}
\end{algorithmic}
\caption{\label{ADAPT}The Adapt algorithm}
\end{algorithm}

\section{Graph Coloring as a Monte Carlo Search Problem}
\label{sec:graph-coloring-as-mc}

In this section we discuss several alternative models to capture the Graph Coloring problem as a Monte Carlo Search problem. Remark that we focus \textit{decision} problem (i.e. deciding if a graph can be colored with a given number of colors). 

We start by defining the possible moves, and then present the node ordering heuristic, and deal with the question of generating valid moves.
Finally, we discuss the objective function and as well as an optimization of the adapt function for the graph coloring problem.

\subsection{Legal Moves}

In the context of the graph coloring problem, a {\em move} consists in assigning a particular color to an uncolored vertex of the graph. Thus, given a graph $G = (V, E)$ and a set of colors $C$, a move is a pair $(v, c)$ where $v$ is an uncolored vertex in $V$, and $c$ is any color in $c$.

NMCS and NRPA only consider {\em legal moves} at each step, and lowering the number of legal moves is a key performance issue for both algorithms. In NMCS, a large number of legal moves leads to a very large branching factor which slows down the algorithm at each recursive level. For NRPA a large number of legal moves results in a large policy vector, which makes it more difficult to train with comparatively less training examples. 

A first naive approach consists in considering every possible move at every step, leading to a number of possible moves that can be as big as $|V| \times |C|$ in the initial condition. This solution results in poor efficiency and low quality solution which we do not report here. To lower the number of possible moves down, we adopt a different model in which each move vertex is considered in a particular order (e.g. random order). 
At each step, only one node and all its legal coloration are considered. 
This reduces the maximum number of moves from $|V| \times |C|$ to $|C|$. As we will see, it leads to good results in practice. 
However, imposing an order over the vertices induces a strong bias over the exploration of the search space, which we study in the next section.

\subsection{Node order}

The naive approach to node ordering is to fix a predefined or random order of the nodes and to color them in this order. A better heuristic is DSatur \cite{brelaz1979new}. It chooses as the next node to color the node that has the less possible colors. In case multiple nodes have the same minimal number of possible colors it break ties by choosing the node that has the most neighbors. DSatur is a good heuristic to order nodes for NRPA since it propagates the constraints in the graph and avoids choosing colors for a node that would reveal inconsistent later due to more constrained neighbors. For example if a node has only one possible color it will always be chosen first by DSatur. By doing so the neighboring nodes have one less possible color and it avoids taking this impossible colors for neighboring nodes which would not have been the case if the one color node had been chosen later.

\subsection{Selective Search}

When trying possible colors for a node it is not wise to choose a color that is already assigned to a neighboring node. In order to avoid as much as possible  bad branching decisions we use forward checking. When selecting the color for a node, all the colors of the neighboring nodes are removed from the set of possible colors for the node. This is related to selective NRPA \cite{cazenave2016selective} where heuristics are used to avoid bad moves. However in our case of Graph Coloring the moves that are discarded are moves that can never be part of a valid solution. So it is safe to remove them. Inconsistent colors are never considered as possible moves except if there a no possible color for a node since neighboring nodes already contains all the available colors. In this case all colors are considered possible and the algorithm chooses a color for the node even if it is inconsistent.

\subsection{Scoring function}
Another design choice is the way to score a playout. 
In~\cite{CazenaveIJCAI09} the depth of the playout was used for Sudoku and the playouts were stopped when reaching an inconsistent state, i.e. a state where a variable has no more possible values. 
For Graph Coloring we use a more informed score. 
We count the number of inconsistent edges, i.e. monochromatic edges. 
If there are two adjacent vertices with the same color, the score decrease by one.
A score equal to the number of edges of the graph means that we have found a solution. 
Note that we also tried a scoring function mixing both the number of colors and the number of inconsistent edges (trying to decrease both), which would allow the algorithm to solve the optimization problem directly, but it didn't work well.

We also experimented with NMCS. For the sake of completeness NMCS is given in algorithm \ref{NMCS}. The score of the playouts, the selection of edges, and the selectivity of colors in NMCS are the same as in NRPA.

\subsection{Coding the moves}

In NRPA it is important to design how moves are coded. 
There is a bijection between moves and integer such that moves are associated to weights. 
We choose to use a simple coding for our moves: the index of a node multiplied by the number of colors plus the index of the color in the move.

\subsection{Adapt all the colors}

When modifying the weights with the adapt function, there are two options. The first one is to modify the weights of the possible moves and to adapt using only the probabilities of the moves that can be played in the current state. The second one is to modify the weights for all the colors, including the colors that were discarded as possible moves since they were inconsistent with neighboring nodes.

The standard Adapt function is given in algorithm~\ref{ADAPT}. The modified function that modifies the probabilities for all the colors is given in algorithm~\ref{ADAPTALL}.

\begin{algorithm}
\begin{algorithmic}
\STATE{AdaptAll ($policy$, $sequence$)}
\STATE{$polp \leftarrow$ $policy$}
\STATE{$state \leftarrow$ $root$}
\FOR{$move$ in $sequence$}
\STATE{$polp$ [code($move$)] $\leftarrow$ $polp$ [code($move$)] + $\alpha$}
\STATE{$z$ $\leftarrow$ 0.0}
\FOR{$m$ in all possible colors even the illegal ones}
\STATE{$z$ $\leftarrow$ $z$ + exp ($policy$ [code($m$)])}
\ENDFOR
\FOR{$m$ in all possible colors even the illegal ones}
\STATE{$polp$ [code($m$)] $\leftarrow$ $polp$ [code($m$)] - $\alpha * \frac{exp (policy [code(m)])}{z}$}
\ENDFOR
\STATE{$state$ $\leftarrow$ play ($state$, $move$)}
\ENDFOR
\STATE{$policy$ $\leftarrow$ $polp$}
\end{algorithmic}
\caption{\label{ADAPTALL}The AdaptAll algorithm}
\end{algorithm}

\section{Compared approaches for graph coloring}
\label{sec:other-approaches-for-graph-coloring}

In this section we present the other Graph Coloring algorithms that we used to compare with our approach.

\subsection{SAT}

We used the following SAT encoding to decide if one can color a graph $G=(V,E)$ with $k$ colors (this formulation is for example used in~\cite{DBLP:conf/ijcai/IgnatievMM17}). 
We add a variable $x_{v,i}$ for each $v\in V$ and each $i \in [k]$.
Then, for each $v \in V$, we add a clause $(\bigvee_i x_{v,i})$, ensuring that each vertex receives a color, and for each edge $uv \in E$ and each color $i \in [k]$, we add a clause $(\neg x_{u,i} \vee \neg x_{v,i})$ such that adjacent vertices receives different colors.
This formula is true iff there is a $k$-coloration of $G$.
Note that if there is no truth assignment for the formula, it tells that the graph is not $k$-colorable.
However, we will not use this in our experiments.

As a solver, we used MiniSat to solve the built formula~\cite{DBLP:conf/sat/EenS03}.

\subsection{HEAD}

HEAD~\cite{moalic2018variations} is an hybrid meta-heuristic, more precisely a memetic algorithm, mixing a local search procedure (Tabu-Search)  with an evolutionary algorithm. It is based on the {\em Hybrid Evolutionary Algorithm} (HEA) by Galinier and Hao~\cite{galinier1999hybrid}. 

The general principle of HEA and HEAD is to start with a population of individuals, which are first improved using a local search procedure.  Then, a crossover operator is applied to the best individuals in order to generate new diverse individuals and the procedure is repeated for a fixed number of steps which are called generations. 

However general crossover operators do not work well for the graph coloring problem, so the main contribution of HEA is a specialized crossover operator. In HEA each individual is a partition of vertices into color classes, and the crossover operator is required to preserve color classes or subset of color classes.

HEAD builds on HEA by introducing various improvements including an original method to maintain diversity inside the population: individuals from earlier generations are re-introduced as candidate individuals in later generations. Using these technique, HEAD has been able to rediscover known coloring at much lower computational cost that earlier approaches \cite{moalic2018variations}.

The source-code of HEAD is available online\footnote{\url{https://github.com/graphcoloring/HEAD/}}.
Authors provided experiments showing good performances on the classical benchmarks and is probably the faster heuristic to date.

\subsection{Greedy Coloring}

For greedy coloring we use the same generation of possible moves as NRPA except that we only play one playout and that the maximum numbers of colors is not fixed. 
The order in which vertices are visited during this greedy coloring is the one given by DSatur~\cite{brelaz1979new}.
Greedy Coloring is used to establish initial upper bounds for NRPA, NMCS, SAT and HEAD that are lowered down using search to establish better upper bounds.

\section{Experiments}
\label{sec:exp}

In this section, we compare the performance of two Monte-Carlo based approaches described in Section~\ref{NRPA} and \ref{sec:graph-coloring-as-mc} with the other approaches described in Section~\ref{sec:other-approaches-for-graph-coloring}.

\subsection{Experimental Protocol}

\paragraph{Execution strategy:}

In practice we observe a high variance in runtimes and best results throughout the different runs of the same algorithm. To reduce the variance in the results, and allow a fair comparison, we proceed as follows: each algorithm is executed 5 times for a given number of color $k$ with a timeout of 30 minutes. If the algorithm discovers at least one valid $k$-coloring for the graph instance, we decrease $k$ by one, and repeat this procedure until the algorithm is unable to discover a $k$-coloring within the timeout limit. Then, we report the lowest $k$ for which the algorithm was able to discover a coloring (denoted {\bf UB} in the result tables), as well as the success rate for this lowest $k$ (denoted {\bf Reached}). The initial value of $k$ to start with is determined with the simple greedy algorithm with nodes ordered according to DSatur (denoted {\bf UBI}). Sometimes the algorithm is unable to improve over the simple greedy algorithm, which we signal with a '--' in the result table (unless the greedy algorithm has already discovered the minimum number $\chi$).

\paragraph{Test instances:}
We used standard benchmark instances available on the website maintained by Gualandi and Chiarandini~\cite{website}, collected from DIMACS benchmark. This benchmark has been used extensively to evaluate graph coloring algorithms and is now considered to be the standard set for experimentation~\cite{DBLP:journals/itor/MalagutiT10} in this field. Moreover, because these instances have been extensively studied, the optimal chromatic number $\chi$ is known for most of them. 

These instances are sorted by difficulty. Instances marked NP-m (resp. NP-d) should be solved in less than a hour (resp. than a day).
For the harder instances marked NP-?, either the chromatic number is unknown or the time needed to solve them is unknown to~\cite{website}.

\paragraph{Hardware and implementations details:}

Every execution reported in this experimental section has been conducted on a Intel Xeon E5-2630 v3 (Haswell, 2.40GHz). %
Although some algorithm support parallel execution, (e.g. NRPA and HEAD), we on report execution times on sequential execution (using one thread) to reduce the variance in execution times, and to allow meaningful comparison with other purely sequential algorithms. 

The peak memory usage is limited to 4GB which is not a limitation for any of the algorithm except for the SAT model which runs out of memory sometimes.

We implemented algorithms using SAT, NRPA and NMCS in C++, using the Boost Graph Library.
For NRPA, we use our implementation described in \cite{negrevergne2017distributed} and available online\footnote{\url{https://github.com/bnegreve/nrpa}}. %
In our experiments we use $l=7, N=100$ for NRPA\footnote{Note that level $7$ will probably never be reached in a reasonable amount of time, this is to allow NRPA to continue to search until a solution is found ; this value of $N$ gave often better results than smaller or bigger values}.
For NMCS, we use increasing nesting levels. 
This mean that we start the search with a level of 1, and if no solution is found, we increment the level and repeat.

\subsection{Results}

\renewcommand{\b}[1]{\textbf{#1}}

\begin{table*}[!ht]
  \centering
    \caption{Results for the easy instances (marked NP-m) with a timeout of 30 minutes.}
  \label{tableNPm}
\begin{tabular}{c|cccc|cc|cc|cc|cc}
  \multicolumn{5}{c|}{} & \multicolumn{2}{c|}{\b{NMCS}} & \multicolumn{2}{c|}{\b{NRPA}} & \multicolumn{2}{c|}{\b{SAT}} & \multicolumn{2}{c}{\b{HEAD}}                                                                                                                                \\
  
\b{Instance}   & $|V|$ & $|E|$ & \b{$\chi$} & \b{UBI} & \b{UB}                     & \b{Reached}                & \b{UB}                     & \b{Reached} & \b{UB} & \b{Reached} & \b{UB} & \b{Reached} \\\hline
1-FullIns\_4   & 93    & 593   & 5          & 5   & \b{5 }    & 100\%    & \b{5}  & 100\%    & \b{5}  & 100\%   & \b{5}  & 100\%    \\
2-FullIns\_4   & 212    & 1621   & 6          & 6   & \b{6 }    & 100\%    & \b{6 } & 100\%    & \b{6 } & 100\%   & \b{6 } & 100\%    \\    
3-FullIns\_3   & 80    & 346   & 6          & 6   & \b{6 }    & 100\%    & \b{6 } & 100\%    & \b{6 } & 100\%   & \b{6 } & 100\%    \\    
4-FullIns\_3   & 114   & 541   & 7          & 7   & \b{7 }    & 100\%    & \b{7 } & 100\%    & \b{7 } & 100\%   & \b{7 } & 100\%    \\    
5-FullIns\_3   & 154   & 792   & 8          & 8   & \b{8 }    & 100\%    & \b{8 } & 100\%    & \b{8 } & 100\%   & \b{8 } & 100\%    \\    
ash608GPIA     & 1216  & 7844  & 4          & 6   & \b{4 }    & 100\% & \b{4 } & 100\% & \b{4 } & 100\%& \b{4 } & 100\% \\           
ash958GPIA     & 1916  & 12506 & 4          & 6   & \b{4 }    & 60\%  & \b{4 } & 100\% & \b{4 } & 100\%& \b{4 } & 100\% \\    
le450\_15a     & 450   & 8168  & 15         & 17  & \b{15}    & 60\%  & \b{15} & 100\% & \b{15} & 100\%& \b{15} & 100\% \\    
mug100\_1      & 100   & 166   & 4          & 4   & \b{4 }    & 100\%    & \b{4 } & 100\%    & \b{4 } & 100\%   & \b{4 } & 100\%    \\    
mug100\_25     & 100   & 166   & 4          & 4   & \b{4 }    & 100\%    & \b{4 } & 100\%    & \b{4 } & 100\%   & \b{4 } & 100\%    \\    
qg.order40     & 1600  & 62400 & 40         & 42  & \b{40}    & 100\% & \b{40} & 100\% & \b{40} & 100\%& \b{40} & 100\% \\    
wap05a         & 905   & 43081 & 50         & 50  & \b{50}    & 100\%    & \b{50} & 100\%    & \b{50} & 100\%   & \b{50} & 100\%    \\    
myciel6        & 95    & 755   & 7          & 7   & \b{7 }    & 100\%    & \b{7 } & 100\%    & \b{7 } & 100\%   & \b{7 } & 100\%    \\    
school1\_nsh   & 352   & 14612 & 14         & 26  & \b{14}    & 100\% & \b{14} & 100\% & \b{14} & 100\%& \b{14} & 100\% \\\hline
\multicolumn{5}{l|}{Avg. ratio to $\chi$}      & \multicolumn{2}{c|}{1.0000} & \multicolumn{2}{c|}{1.0000} & \multicolumn{2}{c|}{1.0000} & \multicolumn{2}{c}{1.0000}                                \\
\end{tabular}

\end{table*}

\begin{table*}[!ht]
  \centering
      \caption{Results for the instances marked NP-h by~\cite{website} with a timeout of 30 minutes.}
  \label{tableNPh}
  \begin{tabular}{c|cccc|cc|cc|cc|cc}
                    \multicolumn{5}{c|}{} & \multicolumn{2}{c|}{\b{NMCS}} & \multicolumn{2}{c|}{\b{NRPA}} & \multicolumn{2}{c|}{\b{SAT}} & \multicolumn{2}{c}{\b{HEAD}}                                                                                                                                \\
    \b{Instance}                          & $|V|$                         & $|E|$                         & $\chi$                       & {\b UBI} & \b{UB}                     & \b{Reached}                & \b{UB}                     & \b{Reached} & \b{UB} & \b{Reached} & \b{UB} & \b{Reached} \\\hline
flat300\_28\_0                            & 300                           & 21695                         & 28                           & 41       &  38                           & 20\%                       &  35                                & 20\%        &  39         & 100\%       & \b{31}     & 100\%       \\ 			  
r1000.5                                   & 1000                          & 238267                        & 234                          & 248      &  243                          & 20\%                       & \b{240}                            & 40\%        &  247        & 100\%       &  248       & --        \\ 			  
r250.5                                    & 250                           & 14849                         & 65                           & 67       & \b{65}                        & 100\%                      & \b{65}                             & 100\%       & \b{65}      & 100\%       & 66     & 40\%        \\ 		  
DSJR500.5                                 & 500                           & 58862                         & 122                          & 132      &  125                          & 60\%                       & \b{122}                            & 40\%        &  126        & 100\%       & 124    & 60\%        \\ 		  
DSJR500.1c                                & 500                           & 121275                        & 85                           & 88       &  88                           & --                         &  87                                & 60\%        & \b{86}      & 100\%       & \b{86}     & 80\%        \\ 		  
DSJC125.5                                 & 125                           & 3891                          & 17                           & 23       &  19                           & 100\%                      &  18                                & 100\%       &  19         & 100\%       & \b{17}     & 100\%       \\ 			  
DSJC125.9                                 & 125                           & 6961                          & 44                           & 50       &  45                           & 40\%                       & \b{44}                             & 100\%       &  46         & 100\%       & \b{44}     & 100\%       \\ 			  
DSJC250.9                                 & 250                           & 27897                         & 72                           & 90       &  84                           & 20\%                       &  76                                & 20\%        &  86         & 100\%       & \b{72}     & 100\%       \\ 			  
queen10\_10                               & 100                           & 2940                          & 11                           & 14       & \b{11}                        & 60\%                       & \b{11}                             & 40\%        &  12         & 100\%       & \b{11}     & 100\%       \\ 			  
queen11\_11                               & 121                           & 3960                          & 11                           & 14       &  13                           & 100\%                      &  13                                & 100\%       &  13         & 100\%       & \b{12}     & 100\%       \\ 			  
queen12\_12                               & 144                           & 5192                          & 12                           & 16       &  14                           & 100\%                      &  14                                & 100\%       &  14         & 100\%       & \b{13}     & 100\%       \\ 			  
queen13\_13                               & 169                           & 6656                          & 13                           & 17       &  15                           & 100\%                      &  15                                & 100\%       &  15         & 100\%       & \b{14}     & 100\%       \\ 			  
queen14\_14                               & 196                           & 4186                          & 14                           & 19       &  16                           & 100\%                      &  16                                & 100\%       &  16         & 100\%       & \b{15}     & 100\%       \\ 			  
queen15\_15                               & 225                           & 5180                          & 15                           & 20       &  17                           & 40\%                       &  17                                & 100\%       &  18         & 100\%       & \b{16}     & 100\%       \\\hline
\multicolumn{5}{l|}{Avg. ratio to $\chi$}      & \multicolumn{2}{c|}{1.1101} & \multicolumn{2}{c|}{1.0851} & \multicolumn{2}{c|}{1.1276} & \multicolumn{2}{c}{1.0428}                                \\
  \end{tabular}

\end{table*}

\begin{table*}[!ht]
  \centering
    \caption{Results for the difficult instances marked NP-? by~\cite{website} with a timeout of 30 minutes.}
  \label{tableNPx}
  \begin{tabular}{c|cccc|cc|cc|cc|cc}
                    \multicolumn{5}{c|}{}                  & \multicolumn{2}{c|}{\b{NMCS}} & \multicolumn{2}{c|}{\b{NRPA}} & \multicolumn{2}{c|}{\b{SAT}} & \multicolumn{2}{c}{\b{HEAD}}                                                                                                                \\
    \b{Instance} & $|V|$                        & $|E|$                        & \b{$\chi$}                  & \b{UBI} & \b{UB} & \b{Reached} & \b{UB}   & \b{Reached} & \b{UB} & \b{Reached} & \b{UB}  & \b{Reached}  \\\hline
le450\_5a        & 450                          & 5714                         & 5                           & 10      & 6      & 20\%   &\b{5}  & 100\% & \b{5} & 100\%   & \b{5}   & 100\%      \\ 			  
le450\_5b        & 450                          & 5734                         & 5                           & 7       & 6      & 40\%   &\b{5}  & 20\%  & \b{5} & 100\%   & \b{5}   & 100\%      \\ 			  
le450\_15b       & 450                          & 8169                         & 15                          & 17      &\b{15}  & 100\%  &\b{15} & 100\% & \b{15}& 100\%   & \b{15}  & 100\%      \\ 			  
le450\_15c       & 450                          & 16680                        & 15                          & 24      & 22     & 100\%  &21     & 100\% &  22   & 100\%   & \b{15}  & 100\%      \\ 		  
le450\_15d       & 450                          & 16750                        & 15                          & 24      & 22     & 100\%  &20     & 20\%  &  22   & 100\%   & \b{15}  & 100\%      \\ 		  
le450\_25c       & 450                          & 17343                        & 25                          & 28      & 27     & 100\%  &\b{26} & 100\% &  27   & 100\%   & \b{26}  & 100\%      \\ 			  
le450\_25d       & 450                          & 17425                        & 25                          & 29      & 27     & 100\%  &\b{26} & 100\% &  27   & 100\%   & \b{26}  & 100\%      \\ 			  
qg.order60       & 3600                         & 212400                       & 60                          & 63      &\b{60}  & 40\%   &62     & 100\% &  61   & 100\%   & \b{60}  & 100\%      \\ 			  
qg.order100      & 10000                        & 990000                       & 100                         & 106     & --    & 20\%   &102    & 20\%  &  --  & 20\%    & \b{100} & 100\%       \\\hline 		  
\multicolumn{5}{l|}{Avg. ratio to $\chi$}      & \multicolumn{2}{c|}{1.7626} & \multicolumn{2}{c|}{1.0963} & \multicolumn{2}{c|}{1.1300} & \multicolumn{2}{c}{1.0089}                                \\
  \end{tabular}

\end{table*}

\renewcommand{\b}[1]{\textbf{#1}}
\begin{table*}[!ht]
  \centering
      \caption{Results for the very difficult problems (NP-?) with a timeout of 30 minutes. The chromatic number of these graphs seems unknown and we only know a lower bound via~\cite{website}.}
  \label{tableNPx2}
  \begin{tabular}{c|cccc|cc|cc|cc|cc}
\multicolumn{5}{c}{} & \multicolumn{2}{c|}{\b{NMCS}} & \multicolumn{2}{c|}{\b{NRPA}} & \multicolumn{2}{c|}{\b{SAT}} & \multicolumn{2}{c}{\b{HEAD}}                                                                        \\
    \b{instance}     & $|V|$                         & $|E|$                         & \b{$\chi_{LB}$}              & \b{UBI} & \b{UB} & \b{Reached} & \b{UB} & \b{Reached} & \b{UB} & \b{Reached} & \b{UB} & \b{Reached} \\\hline
DSJC250.1            & 250                           & 3218                          & 4                            & 10      & 9      & 100\%       & \b{8}  & 40\%        &  9       & 100\%       & \b{8}      & 100\%       \\
DSJC250.5            & 250                           & 15668                         & 26                           & 37      & 34     & 100\%       & 32     & 100\%       &  35      & 100\%       & \b{28}     & 100\%       \\
DSJC500.1            & 500                           & 12458                         & 9                            & 16      & 14     & 40\%        & 14     & 100\%       &  15      & 100\%       & \b{12}     & 100\%       \\
DSJC500.5            & 500                           & 62624                         & 43                           & 65      & 62     & 20\%        & 59     & 80\%        &  63      & 100\%       & \b{48}     & 100\%       \\
DSJC500.9            & 500                           & 112437                        & 123                          & 163     & 161    & 20\%        & 148    & 20\%        &  163     & --          & \b{126}    & 100\%       \\
DSJC1000.1           & 1000                          & 49629                         & 10                           & 25      & 25     & --          & 24     & 100\%       &  25      & --          & \b{21}     & 100\%       \\
DSJC1000.5           & 1000                          & 249826                        & 73                           & 114     & 114    & --          & 112    & 40\%        &  114     & --          & \b{83}     & 60\%        \\
DSJC1000.9           & 1000                          & 449449                        & 216                          & 301     & 301    & --          & 299    & 40\%        &  301     & --          & \b{223}    & 20\%        \\
flat1000\_50\_0      & 1000                          & 245000                        & 15                           & 113     & 113    & --          & 111    & 40\%        &  113     & --          & \b{50}     & 100\%       \\
flat1000\_60\_0      & 1000                          & 245830                        & 14                           & 112     & 112    & --          & 112    & --          &  112     & --         & \b{60}     & 100\%       \\
flat1000\_76\_0      & 1000                          & 246708                        & 14                           & 115     & 115    & --          & 110    & 20\%        &  113     & 100\%       & \b{82}     & 80\%        \\
r1000.1c             & 1000                          & 485090                        & 96                           & 107     & 107    & --          & 107    & --          &  105     & 100\%       & \b{98}     & 20\%        \\
abb313GPIA           & 1557                          & 53356                         & 8                            & 11      & 11     & --          & 11     & --          & \b{9}    & 100\%       & \b{9}      & 60\%        \\
latin\_square\_10    & 900                           & 307350                        & 90                           & 129     & 129    & --          & 121    & 20\%        &  129     & --          & \b{103}    & 20\%        \\
wap01a               & 2368                          & 110871                        & 41                           & 47      & 46     & 60\%        & 45     & 40\%        &  43      & 100\%       & \b{42}     & 60\%        \\
wap02a               & 2464                          & 111742                        & 40                           & 46      & 46     & --          & 45     & 100\%       & \b{42}   & 100\%       & \b{42}     & 100\%       \\
wap03a               & 4730                          & 286722                        & 40                           & 57      & 55     & 40\%        & 55     & 100\%       &  46      & 100\%       & \b{44}     & 20\%        \\
wap04a               & 5231                          & 294902                        & 40                           & 46      & 46     & --          & 46     & --         & \b{44}   & 100\%       & \b{44}     & 100\%       \\
wap06a               & 947                           & 43571                         & 40                           & 44      & 42     & 100\%       & \b{41} & 100\%       & \b{41}   & 100\%       & 42     & 60\%        \\
wap07a               & 1809                          & 103368                        & 40                           & 47      & 44     & 80\%        & 43     & 60\%        & \b{42}   & 100\%       & \b{42}     & 100\%       \\
wap08a               & 1870                          & 104176                        & 40                           & 44      & 43     & 20\%        & 43     & 100\%       & \b{42}   & 100\%       & \b{42}     & 100\%       \\
C2000.5              & 2000                          & 999836                        & 99                           & 207     & 207    & --          & 207    & --          &  207     & --          & \b{151}    & 40\%        \\
C4000.5              & 4000                          & 4000268                       & 107                          & 376     & 376    & --          & 376    & --          &  376     & --          & \b{281}    & 20\%        \\\hline
\multicolumn{5}{l|}{Avg. ratio to $\chi$}      & \multicolumn{2}{c|}{2.3746} & \multicolumn{2}{c|}{2.3173} & \multicolumn{2}{c|}{2.3410} & \multicolumn{2}{c}{1.7027}                                \\
\end{tabular}

\end{table*}

We give results of Monte Carlo approaches (NMCS and NRPA) compared to other approaches in tables~\ref{tableNPm},\ref{tableNPh},\ref{tableNPx} and \ref{tableNPx2}.

First, we observe that the simple greedy algorithm is generally able to discover the $\chi$ value for a number of instances in Table~\ref{tableNPm}. When it is not the case, all the approaches we discuss in this paper have been able to tackle these instances, and discover the $\chi$. However, NMCS does not reach 100\% success rate on two of these instances. 

When we look at the other results, we can see that NMCS is generally the weakest algorithm, and obtains the worst (higher) ratio to $\chi$ for the very difficult instances in Table~\ref{tableNPx} and \ref{tableNPx2}, often significantly worse than the ratio for NRPA. Note that with a different representation of the problem, authors of~\cite{edelkamp2017solving} reported better results with NMCS, contrary to our results.
We also observe that the performance gap between the two approaches is small on the medium NP-m instances (Table~\ref{tableNPm}), but large on the most difficult instances (Table~\ref{tableNPx}). This suggests that with our modeling, NRPA is able to acquire better policies along the execution of the algorithm. The benefit of the learned policies over the tree search becomes more visible in long runs on difficult instances. %
This motivates the general idea of introducing learning into Monte-Carlo search in order to improve the quality of the search. 

We can also see that NRPA is generally better than the SAT model, but SAT is surprisingly good for the difficult wap instances in Table \ref{tableNPx2}, unfortunately, we are unable to explain this result and further experiments are needed.  

However, a pair-wise comparison between the upper bounds discovered by NRPA and the ones discovered by HEAD demonstrates that HEAD is better in general.
It is worth mentioning that algorithms such as HEAD include specialized graph coloring operators which have been extensively studied over the past decades. 
In contrast our algorithm is based on a general purpose implementation of NRPA, and only includes the specializations we have described in Section~\ref{sec:graph-coloring-as-mc}.
Nevertheless, NRPA is the second best in all the datasets, and is often as good, or even better than HEAD.

\begin{sidewaysfigure}
    \centering
    \includegraphics[width=\linewidth]{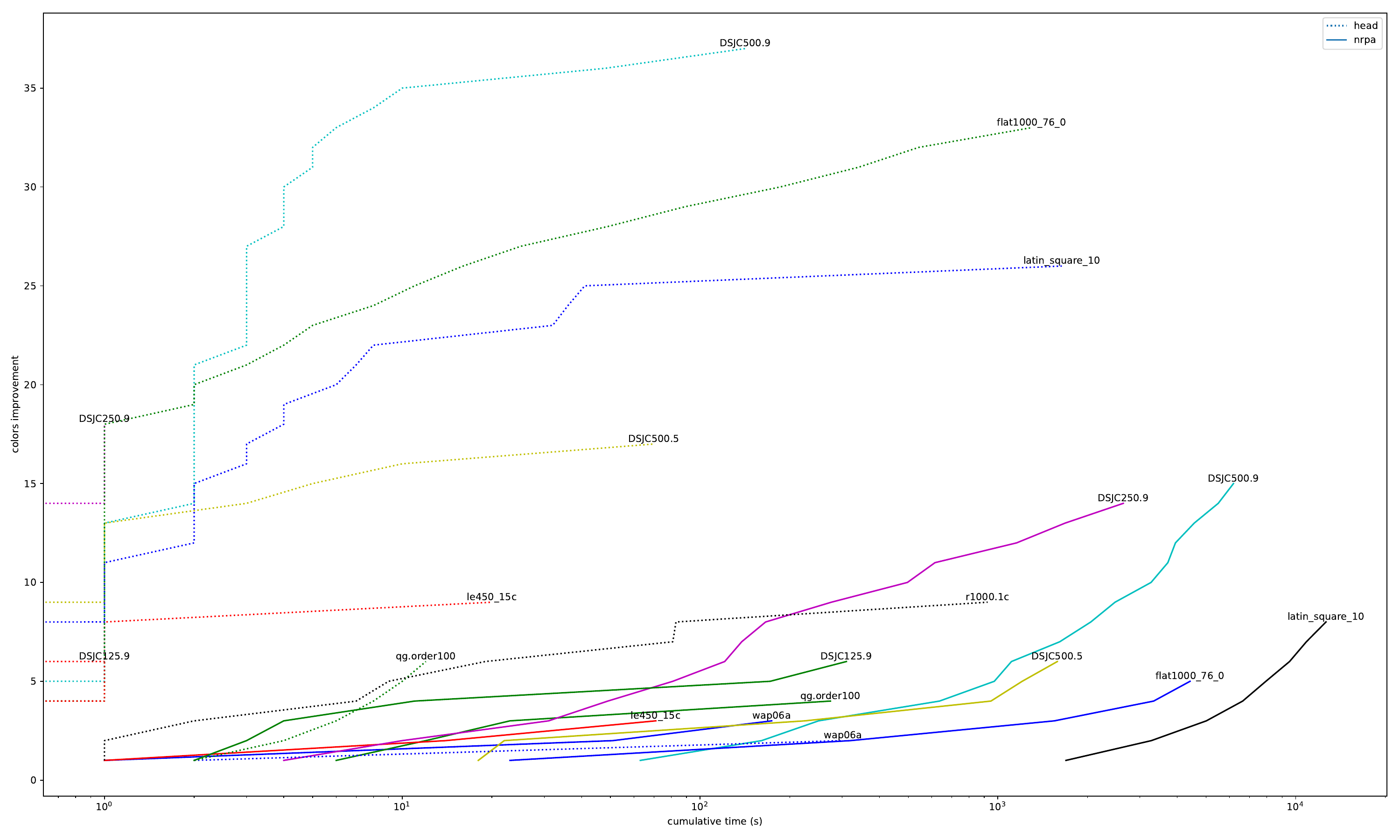}
    \caption{Comparison of computation times (log scale) between NRPA (solid lines) and HEAD (dotted lines) for some instances. }
    \label{fig:plots}
\end{sidewaysfigure}

In Figure~\ref{fig:plots} we further analyze the behaviour of the two algorithms by comparing the execution times. 
We selected the best of 5 runs for each tested number of colors and for each algorithm.
The time is the accumulated time, starting with the number of colors computed by the greedy algorithm.
The y-axis represents the number of improvement by an algorithm for this instance starting from the greedy coloring (i.e. a value of 3 means that the algorithm gave a coloration with 3 less colors than the greedy algorithm did).
For readability, we didn't include instances like C4000.5, where HEAD is very good.
It seems that HEAD improves quite fast the number of colors but struggles to improve more over time (this is not always true however), while NRPA seems to benefit of longer running time.
Indeed, there is not much improvement after 2000s for HEAD while NRPA continues to find colorations after 3000s. 
This could be because HEAD is stuck in local optimum and cannot find new solutions, while NRPA continues to improve policies while exploring the search space.

This demonstrate that Monte Carlo based algorithms have the ability to compete with state-of-the-art hybrid algorithms on the graph coloring problem, and deserve further investigation with more optimizations (more specific strategies, restarts...). From a broader perspective, this also shows that a continuous optimization algorithms can be used to solve discrete problems such as the graph coloring problem.

\section{Conclusion}

In this paper, we deepen our understanding of Monte Carlo Search algorithms applied to Graph Coloring.
Our method is significantly different from most other  methods from the literature, and yet, it is able to compete with state-of-the-art algorithms which have been intensively optimized during the past decades.
These results suggest that Monte Carlo search combined with policy adaptation are able to explore the search to discover good, yet diverse solutions. And that this technique should be investigated further alongside with more standard approaches. 

It would be also interesting to see if NMCS could combine with a good heuristic like HEAD, by either using the local search to improves the solution or by using the genetic algorithm as playouts. 

Future works includes the use of Graph Convolution Networks to model more complex policies that can make branching decisions based on the structure of graph at hand, and generalize knowledge from one graph to another. 

Finally, we could apply Monte Carlo methods to other variants of graph coloring, like for example Weighted Vertex Coloring or Minimum Sum Coloring, since only the evaluation function would change.

\section*{Acknowledgment}

This work was supported in part by the French government under management of Agence Nationale de la Recherche as part of the “Investissements d’avenir” program, reference ANR19-P3IA-0001 (PRAIRIE 3IA Institute).

\bibliographystyle{abbrv}
\bibliography{nrpa}

\end{document}